\documentclass{article}
\usepackage{amsmath}
\usepackage{graphicx}
\usepackage{array}
\usepackage{graphicx}
\usepackage{tabularx}
\usepackage{geometry}
\usepackage{booktabs}
\title{KunLunBaize-VoT-R1: An efficient video inference model integrating image packing and AoE architecture}
\author{Cheng Li, Jiexiong Liu, Yixuan Chen, Yanqin Jia}
\date{}
\begin{document}

\maketitle
\begin{abstract}
In the field of video-language pretraining, existing models face numerous challenges in terms of inference efficiency and multimodal data processing. This paper proposes a KunLunBaize-VoT-R1 video inference model based on a long-sequence image encoder, along with its training and application methods. By integrating image packing technology, the Autonomy-of-Experts (AoE) architecture, and combining the video of Thought (VoT), a large language model (LLM) trained with large-scale reinforcement learning, and multiple training techniques, the efficiency and accuracy of the model in video inference tasks are effectively improved. Experiments show that this model performs outstandingly in multiple tests, providing a new solution for video-language understanding.
\end{abstract}

\section{Introduction}
Multimodal understanding of video and language is a critical area in artificial intelligence, with broad applications in tasks such as video content description, cross-modal retrieval, and spatio-temporal reasoning \cite{r1}. Deep learning theories have laid the foundation for this field, emphasizing the importance of multimodal data processing in intelligent systems. In recent years, large-scale pretraining models like VideoBERT \cite{r2} and Flamingo \cite{r3} have significantly advanced cross-modal tasks by jointly learning visual and textual representations. However, existing methods still face several challenges. The high spatio-temporal complexity of video sequences results in low inference efficiency, making real-time applications difficult \cite{r4, r22}. Additionally, the heterogeneous nature of multimodal data—such as varying image resolutions and text lengths—poses challenges for uniform modeling. Existing methods struggle particularly with integrating fine-grained semantics and large-scale temporal logic\cite{r5}.

Meanwhile, the Chain-of-Thought (CoT) reasoning mechanism has shown great potential in decomposing complex problems and driving multi-step reasoning in text-based scenarios \cite{r6}. By extending this approach to the video domain, the VoT\cite{video-of-thought} reasoning framework inherits the core principles of CoT to break down intricate tasks into simpler, more manageable sub-problems, thus enabling more effective reasoning in video-related tasks. 

However, current multimodal models based on VoT primarily rely on supervised fine-tuning (SFT)\cite{r43}, which lacks VoT-specific training for the core large language models (LLMs). This limitation results in logical inconsistencies and semantic biases during cross-modal reasoning \cite{r7}. Furthermore, although large-scale reinforcement learning has been proven to outperform traditional SFT methods in terms of generalization and robustness \cite{r8}, its integration with multimodal feature extraction remains an underexplored area.

To address these challenges, this study proposes a novel multimodal encoder that integrates image packing technology and the AoE\cite{r41} architecture. Subsequently, the KunLunBaize-VoT-R1 model is constructed through post - training based on reinforcement learning.
Specifically, in the feature extraction stage, an innovative image packing strategy is introduced. Multiple images of different sizes are packed into patches and then input into the visual encoder. A block mask is utilized to isolate the interference between sub-images, and the position encoding is adjusted to guarantee the mathematical equivalence between packed and unpacked sequences\cite{r45}. This approach significantly enhances resource utilization and computational parallelism. Furthermore, an AoE architecture with dense learnable residual connections is adopted. Expert networks are dynamically allocated for sub - images of different scales to achieve the adaptive fusion of fine - grained features and global semantics\cite{r46}. In the inference stage, the encoder is connected to the KunLunBaize-VL model. A structured reward function is employed to guide the model to generate multimodal inference videos that conform to human cognitive logic. Experiments demonstrate that our method achieves leading performance in tasks such as video question answering and temporal action localization\cite{r42}.

\textbf{Our main contributions are as follows:}

\begin{itemize}
    \item Proposing a scalable image packing strategy combined with a block masking mechanism to break through the efficiency bottleneck of multi-image parallel processing.
    \item Design a long-sequence encoder based on the Autonomy-of-Experts  (AoE), dense learnable residual connections, and a hybrid of linear and traditional attention mechanisms to achieve adaptive encoding of heterogeneous modal information. 
    \item For the first time, combining the reinforcement learning post-training paradigm with multimodal VoT reasoning to construct an end-to-end multimodal logical reasoning system. The experimental results verify the dual advantages of the model in terms of efficiency and accuracy.
\end{itemize} 

\section{Related work}
Video-Language Pre-training Models\cite{r25}. In recent years, video-language pre-training models such as VideoBERT and Flamingo have emerged one after another, and have significantly improved the performance of cross-modal tasks through joint modeling of visual and text information. However, when facing the high spatio-temporal complexity of video sequences, their inference efficiency is still greatly limited. In addition, the heterogeneity problems of low-resolution and high-resolution, local and global semantics within multimodal data still remain important factors affecting model performance\cite{r2,r3,r4,r5}.

CoT Reasoning Mechanism. The CoT technology has achieved remarkable results in the pure text field by guiding the model to perform multi-step reasoning. However, when extending CoT reasoning to multimodal scenarios, ensuring the logical coherence and semantic consistency of visual and text information during the reasoning process has become a major challenge. Currently, multimodal CoT models mostly rely on the SFT strategy and lack training specifically for cross-modal reasoning, resulting in the problem of broken reasoning videos in complex tasks\cite{r6,r7}.

AoE and Image Packing Strategies. AoE is a novel MoE (Mixture of Experts)\cite{r27} paradigm in which experts autonomously select themselves to process the input. AoE is based on the insight that experts are aware of their ability to effectively process tokens, and this awareness is reflected in the scale of their internal activations. In AoE, the router is removed; instead, experts pre-compute the internal activations of the input and rank them according to their activation norms. Only the top-ranked experts continue with the forward pass, while the others abort. The overhead of pre-computing activations is reduced through low - rank weight decomposition. This approach of self-evaluation first and then comparison with partners ensures improved expert selection and efficient learning.  At the same time, as a means to improve parallel computing efficiency, the image packing strategy synthesizes multiple low-resolution sub-images into high-resolution images, effectively reducing the computational complexity. With the help of block masking and position encoding adjustment, the mutual independence of each sub-image is ensured. Although these two technologies have been successfully applied in their respective fields, how to organically integrate them to address the heterogeneity challenges of multimodal data in video sequences has not yet formed a systematic study\cite{r9,r10}.

Application of Reinforcement Learning in Multimodal Reasoning. Reinforcement learning technology has been widely used in recent years to improve the generalization and robustness of models, especially in complex reasoning tasks. Compared with traditional supervised fine-tuning methods, models trained based on large - scale reinforcement learning are more capable of generating reasoning videos that conform to human cognitive logic. However, how to combine reinforcement learning with efficient multimodal feature extraction mechanisms to construct an end-to-end multimodal reasoning system is still an important frontier of current research\cite{r8,r11}.

Although the above-mentioned technologies have solved some problems in multimodal reasoning to a certain extent, there are still great challenges in achieving unified modeling of the complex spatio temporal structure and multimodal data heterogeneity features in video sequences while ensuring efficient inference. Against this background, this paper proposes the KunLunBaize-VoT-R1 model, aiming to construct an end-to-end multimodal logical reasoning system through the deep integration of image packing and AoE architecture, combined with a reinforcement learning training strategy, providing new ideas and methods for video-language understanding\cite{r47,r48}.

\begin{figure}[htbp]
\centering
\includegraphics[width=0.8\textwidth]{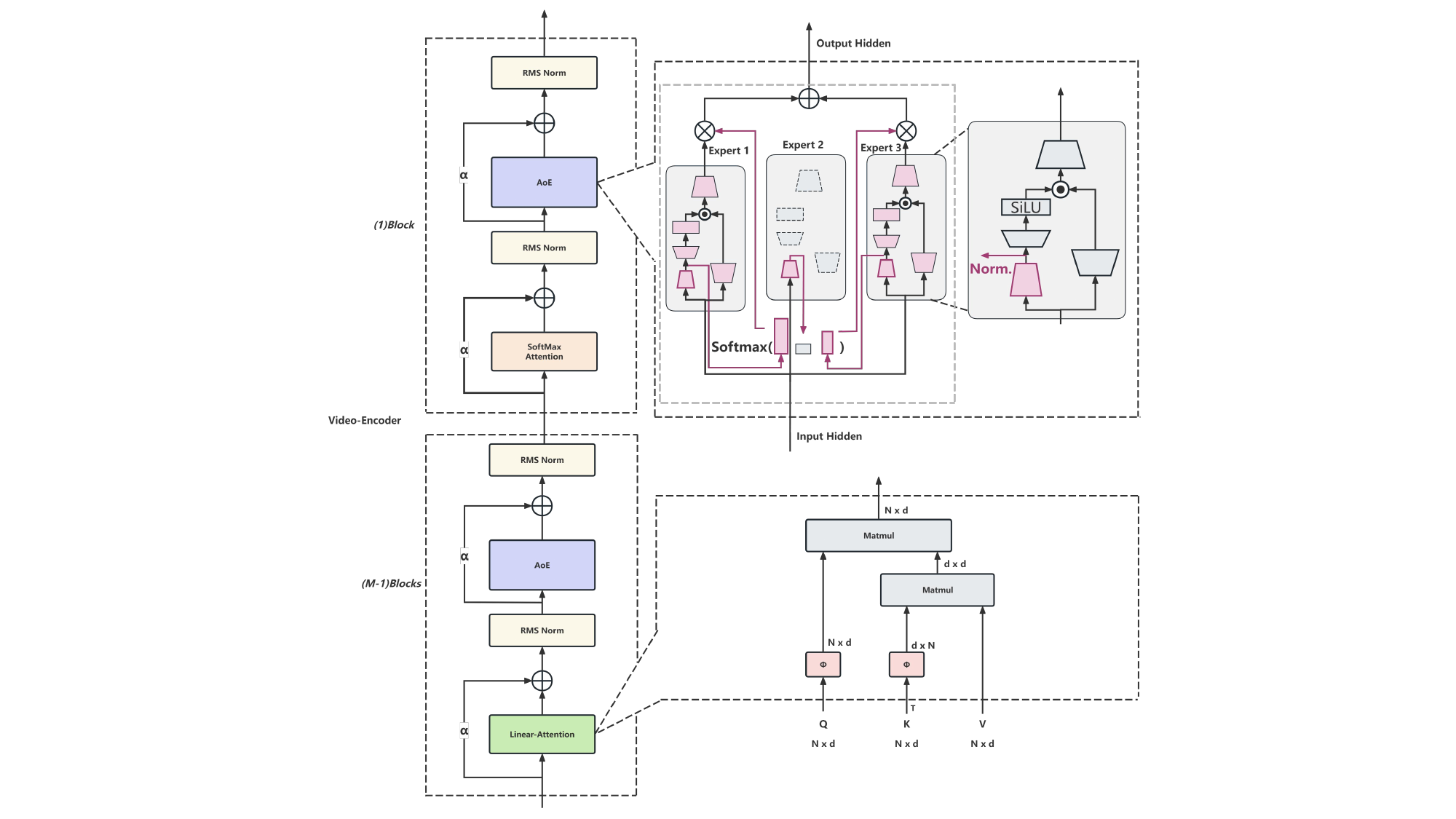}
\caption{The overall architecture of the proposed encoder demonstrates the integration of key components, including the hybrid attention mechanism, the Autonomy-of-Experts (AoE) model, dense learnable residual connections, and sample packing technology.}
\label{fig:example1}
\end{figure}

\begin{figure}[htbp]
\centering
\includegraphics[width=0.8\textwidth]{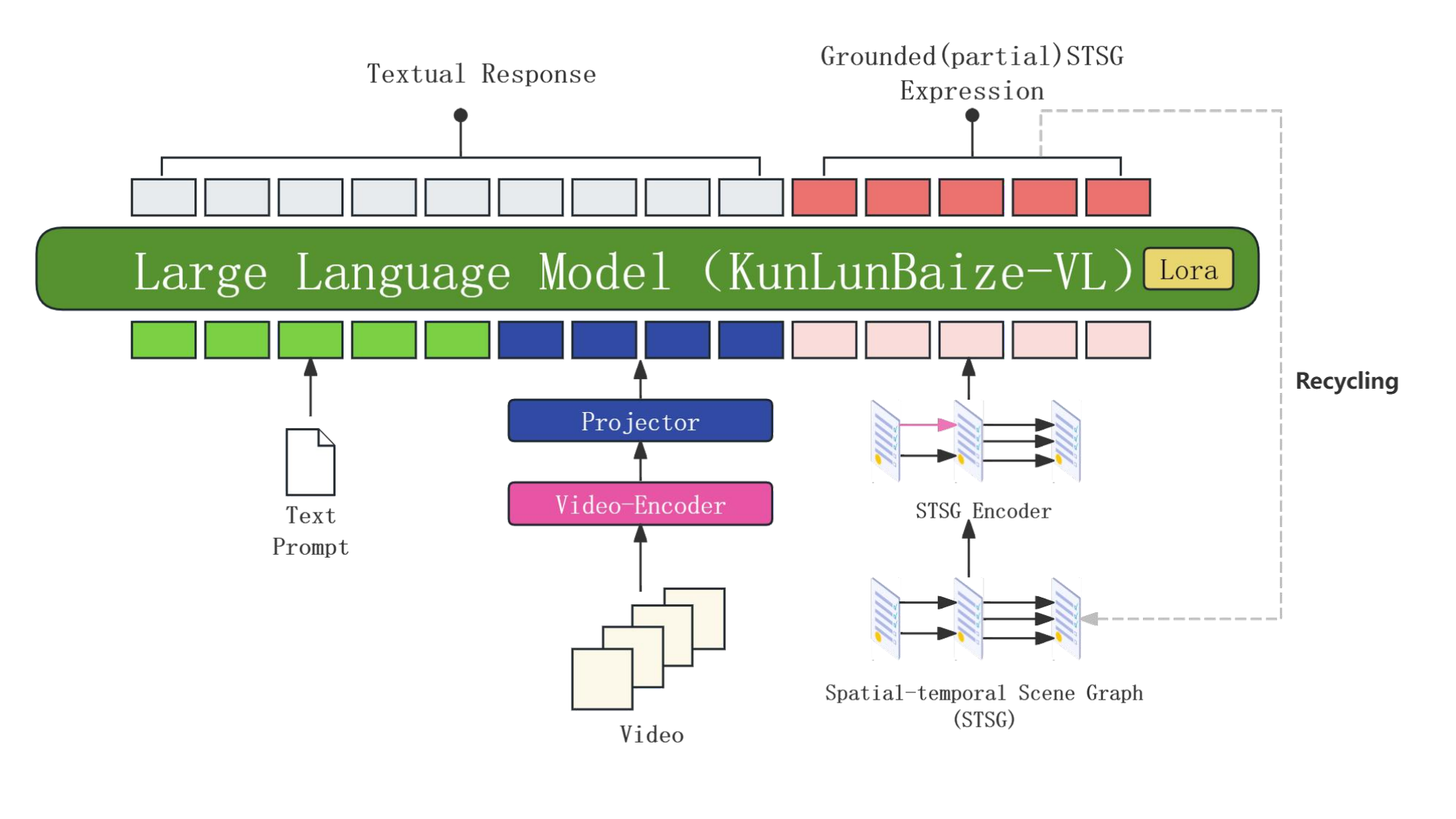}
\caption{VoT reasoning framework}
\label{fig:example2}
\end{figure}

\section{Method}

In this section, we elaborate in detail on the proposed method for long-sequence image and video processing. We introduce some key components, such as the hybrid attention mechanism, the Autonomy-of-Experts (AoE) model, dense learnable residual connections, and sample packing technology. These components jointly enhance the efficiency and performance of processing. The training process includes knowledge distillation, contrastive learning, and multi-stage reinforcement learning strategies to optimize the model for complex reasoning tasks. We follow the Video-of-Thought video reasoning framework, which is illustrated in Figure \ref{fig:example2}.

\subsection{Construction of Long-Sequence Image Encoder}

\subsubsection{Hybrid Attention Mechanism}
To efficiently process long sequences, we adopt an image encoder based on the hybrid attention mechanism, which integrates linear attention and traditional Softmax Attention. The computational complexity of linear attention is $O(d^2 \times L)$, where $d$ is the feature dimension and $L$ is the sequence length. This is significantly more efficient than the $O(L^2 \times d)$ complexity of traditional Softmax Attention. However, relying solely on linear attention may lead to performance degradation. Therefore, we add a layer of conventional Softmax Attention after multiple layers of linear attention to mitigate this issue. The calculation formulas for linear attention and Softmax Attention are as follows:

\begin{equation}
A_{\text{linear}} = \frac{1}{C} \sum_{j=1}^{L} K_{ij} V_j
\end{equation}
where $C$ is the normalization constant, $K_{ij}$ is the similarity between the $i$-th query and the $j$-th key, and $V_j$ is the $j$-th value. The final output is obtained by multiplying the query matrix $Q$ with the linear attention result, i.e., $A_{\text{final}} = Q \cdot A_{\text{linear}}$.

\begin{equation}
A_{\text{softmax}} = \text{softmax}\left(\frac{Q K^T}{\sqrt{d}}\right) V
\end{equation}
where $Q$, $K$, and $V$ are the query, key, and value matrices, respectively.

\subsubsection{Autonomy-of-Experts (AoE)}
The AoE model addresses limitations in traditional Mixture-of-Experts (MoE) models by dynamically selecting the most relevant experts based on the input. The detailed calculation process is as follows:

\textbf{Expert Computation:} For the $i$-th expert, the output is computed as:
\begin{equation}
E_i(x) = \left(\text{SiLU}\left(x W_{\text{down}}^i W_{\text{up}}^i\right) \odot \left(x W_p^i\right)\right) W_o^i
\end{equation}
Here, $x \in \mathbf{R}^{d_{\text{model}}}$ is the input hidden state. $W_{\text{down}}^i \in \mathbf{R}^{d_{\text{model}} \times d_{\text{low}}}$ and $W_{\text{up}}^i \in \mathbf{R}^{d_{\text{low}} \times d_{\text{model}}}$ are matrices obtained from the low-rank factorization of the original weight matrix. $W_p^i \in \mathbf{R}^{d_{\text{model}} \times d_{\text{ffn}}}$ and $W_o^i \in \mathbf{R}^{d_{\text{ffn}} \times d_{\text{model}}}$ are expert-specific weight matrices.

\textbf{Activation Cache Computation:} To improve efficiency, the matrices $W_{\text{down}}^i$ of all experts are combined into a large matrix:
\begin{equation}
\hat{W}_{\text{down}} = \left[W_{\text{down}}^1, \cdots, W_{\text{down}}^n\right] \in \mathbf{R}^{d_{\text{model}} \times (n d_{\text{low}})}
\end{equation}
The activation cache $C$ is then computed as:
\begin{equation}
C = x \hat{W}_{\text{down}}
\end{equation}
The resulting $C \in \mathbf{R}^{n d_{\text{low}}}$ is reshaped into an $n \times d_{\text{low}}$ matrix for subsequent computations.

\textbf{Activation Norm Calculation and Expert Selection:} The $L^2$ norm of the activation cache $C$ is calculated to measure the activation strength of each expert:
\begin{equation}
p = L2\text{-Norm}(C, \text{dim}=-1)
\end{equation}
This yields a vector $p \in \mathbf{R}^n$ containing the activation norm values of each expert. The top-$K$ experts are selected based on these values:
\begin{equation}
I = \text{argtopK}(p)
\end{equation}
The Softmax operation is then applied to the activation norm values of the top-$K$ experts:
\begin{equation}
\hat{p} = \text{softmax}(p[I])
\end{equation}
This results in a probability distribution vector $\hat{p} \in \mathbf{R}^K$ used for weighted calculations.

\textbf{Output Computation:} Only the top-$K$ experts continue the computation. The final output $h$ is calculated as:
\begin{equation}
h += \hat{p}[i] \cdot \left(\text{SiLU}(C[i] W_{\text{up}}^i) \odot (x W_p^i)\right) W_o^i
\end{equation}
For each expert, the output is weighted by $\hat{p}[i]$ and accumulated into the final output $h$.

\subsubsection{Dense Learnable Residual Connection}
To enhance the model's ability to capture complex hierarchical feature relationships, we incorporate dense learnable residual connections. Suppose the output of the $l$-th layer is $H_l$, the connection can be expressed as:
\begin{equation}
H_{l+1} = \text{Layer}(H_l) + \sum_{i=0}^{l} \alpha_i \cdot H_i
\end{equation}
where $\alpha_i$ is a learnable parameter.

\subsubsection{Sample Packing Technology}
To improve pre-training efficiency, we employ greedy packing to combine multiple image patches and input them into the model. A block mask mechanism and adjusted position encoding are used to ensure independent calculations for different images. Suppose the set of packed image patches is $\{P_1, P_2, \cdots, P_n\}$, the position encoding matrix is $E_{\text{pos}}$, and the block mask matrix is $M$, the input data is:
\begin{equation}
X = M \cdot (P + E_{\text{pos}})
\end{equation}

\subsubsection{SizeEmbedding}
A SizeEmbedding is assigned to each image and packed with the image patches before inputting into the model. If the image size is $(w, h)$, the SizeEmbedding is expressed as $E_{\text{size}}(w, h)$. The packed input data is:
\begin{equation}
X_{\text{input}} = [P; E_{\text{size}}(w, h)]
\end{equation}
In conjunction with the AoE expert model, different expert modules are called for images of different sizes.

\subsubsection{Image Encoder Training}
The hybrid attention image encoder is obtained through knowledge distillation from a pre-trained large-scale image encoder (OpenClip-large)~\cite{r14}. The distillation loss $L_{\text{distill}}$ is expressed as:
\begin{equation}
L_{\text{distill}} = \alpha \cdot L_{\text{CE}}(y_{\text{student}}, y_{\text{teacher}}) + (1 - \alpha) \cdot \text{MSE}(f_{\text{student}}(x), f_{\text{teacher}}(x))
\end{equation}
where $L_{\text{CE}}$ is the cross-entropy loss, $y_{\text{student}}$ and $y_{\text{teacher}}$ are the prediction results of the student and teacher models, respectively, $\text{MSE}$ is the mean squared error loss, and $\alpha$ is the balance parameter.

Subsequently, large-scale contrastive learning training is conducted using the image packing method. In the first stage, the contrastive learning loss $L_{\text{contrast1}}$ is:
\begin{equation}
L_{\text{contrast1}} = -\sum_{i=1}^{N} \log \frac{\exp(\text{sim}(z_i, z_i^+)/\tau)}{\sum_{j=1}^{2N} \exp(\text{sim}(z_i, z_j)/\tau)}
\end{equation}
where $z_i$ is the feature of the $i$-th sample, $z_i^+$ is its positive sample pair, $\text{sim}$ represents the similarity calculation, and $\tau$ is the temperature parameter~\cite{r15}.

In the second stage, images are randomly scaled, and the contrastive learning loss $L_{\text{contrast2}}$ is defined similarly, with the positive sample pair now consisting of the original image and its scaled version:
\begin{equation}
L_{\text{contrast2}} = -\sum_{i=1}^{N} \log \frac{\exp(\text{sim}(z_i, z_i^+)/\tau)}{\sum_{j=1}^{2N} \exp(\text{sim}(z_i, z_j)/\tau)}
\end{equation}
This stage aims to enforce consistent feature representations across different image sizes.

\subsection{Video Encoder Training}

\subsubsection{Initialization}
The video encoder is initialized with the pre-trained image encoder obtained from the previous stage. This approach significantly reduces training time and accelerates model convergence~\cite{r14}.

\subsubsection{Contrastive Learning Training}
The video encoder is trained using contrastive learning, where images within the video are randomly and uniformly scaled during training. The contrastive learning loss $L_{\text{video-contrast}}$ is defined as:
\begin{equation}
L_{\text{video-contrast}} = -\sum_{t=1}^{T} \sum_{i=1}^{N} \log \frac{\exp(\text{sim}(z_{t,i}, z_{t,i}^+)/\tau)}{\sum_{j=1}^{2N} \exp(\text{sim}(z_{t,i}, z_{t,j})/\tau)}
\end{equation}
where $T$ is the number of video frames, $z_{t,i}$ is the feature of the $i$-th sample at time step $t$, and $z_{t,i}^+$ is its positive sample pair~\cite{r15}. This loss function encourages the model to generate consistent features for the same video content across different scales.

\subsection{Video-of-Thought Reasoning Framework}

To achieve in-depth understanding and reasoning of complex videos, we adopt the Video-of-Thought (VoT) reasoning framework, which breaks down the reasoning process into multiple steps, transitioning from pixel-level perception to cognitive interpretation\cite{r50}. By integrating the Spatial-Temporal Scene Graph (STSG)\cite{r31} representation with a Multimodal Large Language Model (MLLM), the framework enables pixel-level spatiotemporal video grounding and supports complex Video Question Answering (QA) tasks. The reasoning process is outlined as follows:

\begin{enumerate}
    \item \textbf{Pixel-level Perception:} The framework begins with fine-grained pixel-level perception to identify key objects and scenes in the video. This step lays the foundation for subsequent reasoning by accurately locating and recognizing target objects within the video frames using the STSG representation\cite{r32}.
    
    \item \textbf{Spatio-Temporal Tracking:} After identifying key objects, the framework tracks their spatiotemporal trajectories using the STSG. This step not only focuses on the positional changes of objects but also integrates contextual information to understand their behavioral patterns. The dynamic connections in the STSG enable the model to follow the state changes of objects across frames, providing support for semantic understanding\cite{r33}.
    
    \item \textbf{Action Semantic Understanding:} Based on the spatiotemporal tracking results, the framework analyzes the semantics of object actions. By integrating commonsense knowledge, the model interprets the behaviors of objects, such as identifying whether an action implies a specific intention or outcome. The semantic understanding capabilities of the MLLM allow the model to link actions observed in the video with real-world commonsense knowledge, enabling a deeper understanding of the actions.
    
    \item \textbf{Causal Reasoning:} With an understanding of action semantics, the framework performs causal reasoning to analyze the causal relationships between actions. This step aims to understand the logical sequence of events and potential causes within the video, predicting possible outcomes. The MLLM's reasoning capabilities enable the generation of inference videos that conform to human cognitive logic, enhancing the accuracy and explainability of the reasoning.
    
    \item \textbf{Result Verification:} Finally, the framework verifies the reasoning results from both pixel-level perception and commonsense cognition perspectives. This dual verification ensures that the results are consistent with the video content and real-world logic, effectively avoiding incorrect inferences and improving the reliability of the reasoning outcomes.
\end{enumerate}

\subsection{Multi-stage Reinforcement Learning Strategy}

\subsubsection{Adapter-Only Training}
We use KunLunBaize-VL as the large language model and Video-of-Thought as the pre-trained STSG, respectively.  We freeze the LLM, STSG and encoder parts and train only the adapter module through Supervised Fine-Tuning (SFT). Let $\theta_{\text{adapter}}$ denote the adapter parameters. The SFT loss $L_{\text{SFT1}}$ is defined as:
\begin{equation}
L_{\text{SFT1}} = \sum_{i=1}^{M} L_{\text{CE}}(y_{\text{pred}}(x_i; \theta_{\text{adapter}}), y_{\text{true}}(x_i))
\end{equation}
where $x_i$ is the input sample, $y_{\text{pred}}(x_i; \theta_{\text{adapter}})$ is the adapter's prediction, $y_{\text{true}}(x_i)$ is the true label, and $M$ is the number of training samples~\cite{r18}.

\subsubsection{LLM Fine-Tuning with LoRA}
Next, we unfreeze the LLM and fine-tune it using the Low-Rank Adaptation (LoRA) method~\cite{r30,r52}. LoRA approximates the full parameter update through low-rank matrix decomposition. Suppose the original weight matrix is $W$, and the updated weight matrix is $W + \Delta W$, where $\Delta W = BA$, and $B$ and $A$ are low-rank matrices.

\subsubsection{Reinforcement Learning Training for the Base Model}
We follow the training method of DeepSeekR1 to obtain the KunlunBaize-VoT-Zero model through pure reinforcement learning. In reinforcement learning, the agent interacts with the environment, selects an action $a$ according to the policy $\pi$, and receives a reward $r$. The state transition is defined as $s_{t+1} = f(s_t, a_t)$. The goal is to maximize the cumulative reward:
\begin{equation}
R = \sum_{t=0}^{T} \gamma^t r_t
\end{equation}
where $\gamma$ is the discount factor.

\subsubsection{Cold Start and Supervised Fine-Tuning}
We perform a cold start by using thousands of high-quality long VoT samples to conduct SFT on the base model, providing it with initial inference capabilities. The SFT loss $L_{\text{SFT2}}$ at this stage is similar to $L_{\text{SFT1}}$:
\begin{equation}
L_{\text{SFT2}} = \sum_{i=1}^{M} L_{\text{CE}}(y_{\text{pred}}(x_i; \theta_{\text{adapter}}), y_{\text{true}}(x_i))
\end{equation}
This stage aims to equip the model with basic reasoning abilities.

\subsubsection{Enhancing Inference Ability through Reinforcement Learning}
We conduct reinforcement learning to further enhance the model's inference capabilities, using the same large-scale reinforcement learning method as in the Video-VoT-Zero training stage~\cite{r17}. This step focuses on improving the model's ability to generate coherent and accurate reasoning videos.

\subsubsection{Rejection Sampling and Non-Inference Ability Optimization}
We perform rejection sampling and additional SFT to optimize the model's non-inference abilities. In rejection sampling, a threshold $\epsilon$ is set. If the prediction probability $p$ is less than $\epsilon$, the sample is rejected and re-sampled. The SFT loss remains similar to previous stages:
\begin{equation}
L_{\text{SFT3}} = \sum_{i=1}^{M} L_{\text{CE}}(y_{\text{pred}}(x_i; \theta_{\text{adapter}}), y_{\text{true}}(x_i))
\end{equation}
This step ensures that the model performs well in tasks beyond reasoning, such as classification or generation.

\subsubsection{Final Training for Human-Preference Conformity}
We conduct reinforcement learning to align the model's behavior with human preferences, improving its usability and safety. This is achieved by constructing a reward function $R_{\text{human-preference}}$ that incorporates factors related to human preferences, such as rewards for commonsense reasoning and penalties for harmful outputs. The final model obtained is KunLunBaize-VoT-R1, a VoT model capable of video inference while adhering to human preferences.

\section{Experiments}

\subsection{Dataset}

In the initial phase of our experiments, we utilized a dataset comprising 6.22 million image-text pairs sourced from LAION-CC-SBU\cite{r35}. These pairs were captioned using the Qwen2.5-vl\cite{r44} model. The original data was extracted from the CC3M dataset and subsequently filtered through the video-LLaVA\cite{r51} framework to ensure high-quality annotations. For video-text pairs, we obtained 6.84 million pairs from a subset provided by Valley, with the original data derived from the WebVid dataset\cite{r36}.

During the reinforcement learning phase, we compiled an instruction dataset of 86.4K samples from LLaVA-CoT-100k\cite{LLaVA-CoT-100k}. Additionally, we incorporated various video QA datasets, including VLEP\cite{r53}, STAR\cite{r54}, IntentQA\cite{r37}, Social-IQ\cite{r38}, CausalVidQA\cite{r55}, and NExT-QA\cite{r39}, to enrich the training corpus and enhance the model's reasoning capabilities. Finally, we also directly applied reinforcement learning to post-train and test the model on two datasets, DVD-counting-test\cite{DVD-counting-test} and open-r1-video-4k\cite{open-r1-video}. 

\subsection{Model Settings}

For the image encoder, we utilized OpenCLIP-L/14 as the teacher model. We initialized our image encoder through knowledge distillation and then fine-tuned it for our specific tasks. The video encoder was initialized with a pre-trained image encoder to effectively leverage the learned visual features. Regarding the large language model (LLM) component, we chose KunLunBaize-VL. It has powerful multimodal reasoning capabilities and can support complex video inference tasks. 

\subsection{Training Hyperparameter Settings}

During training, we set the learning rate for the image encoder to \(2 \times 10^{-5}\) and utilized the AdamW optimizer to ensure efficient convergence. For the reinforcement learning stage, we designed a reward function that combines accuracy and consistency metrics, guiding the model to generate inference videos that are both accurate and aligned with human cognitive logic.

For the video encoder training, we applied random uniform scaling to video frames within the range of \(0.5\) to \(1.5\) to enhance the model's adaptability to varying video resolutions. We also set the temperature parameter for contrastive learning to \(0.07\) and used a batch size of \(256\) to balance computational efficiency and model performance.

For the large language model (LLM), we employed Low-Rank Adaptation (LoRA) for fine-tuning, decomposing the original weight matrices into low-rank matrices to improve training efficiency and reduce computational overhead. 

\subsection{Main Performance on Video QA Reasoning}

To comprehensively evaluate the reasoning capabilities of our proposed KunLunBaize-VoT-R1 framework in video question answering (QA) tasks, we conducted extensive experiments across multiple benchmark datasets. These datasets are designed to test various reasoning skills, including spatio-temporal understanding, causal reasoning, and commonsense inference. The results demonstrate that our model not only achieves state-of-the-art performance but also significantly outperforms existing methods in terms of accuracy and reasoning depth.

\begin{table}[ht]
\centering
\begin{tabular}{lcccccccc}
\toprule
\textbf{Model} & \multicolumn{1}{c}{\textbf{VLEP}} & \multicolumn{4}{c}{\textbf{STAR}} & \multicolumn{1}{c}{\textbf{IntentQA}} & \multicolumn{2}{c}{\textbf{Social-IQ}} \\
\cmidrule(lr){2-2} \cmidrule(lr){3-6} \cmidrule(lr){7-7} \cmidrule(lr){8-9}
 &  & \textbf{Int.} & \textbf{Seq.} & \textbf{Pre.} & \textbf{Fea.} &  & \textbf{2-Way} & \textbf{4-Way} \\
\midrule
    \multicolumn{9}{l}{\textbullet~\textbf{Baselines}} \\
InternVideo & 63.9 & 62.7 & 65.6 & 54.9 & 51.9 & - & - & - \\
LLAMA-VQA & 71.0 & 66.2 & 67.9 & 57.2 & 52.7 & - & - & - \\
VLAP & 69.6 & 70.0 & 70.4 & 65.9 & 62.2 & - & - & - \\
SeViLA & 68.9 & 63.7 & 70.4 & 63.1 & 62.4 & - & - & - \\
VideoChat & 62.0 & 63.2 & 66.8 & 54.1 & 49.6 & 59.3 & 67.7 & 37.8 \\
Video-LLaVA & 65.7 & 65.0 & 67.7 & 57.8 & 52.0 & 63.2 & 69.5 & 40.4 \\

\midrule
\multicolumn{9}{l}{\textbullet~\textbf{VoT}} \\
MotionEpic & 73.4 & 71.5 & 72.6 & 66.6 & 62.7 & 70.8 & 72.8 & 45.0 \\
KunLunBaize-VoT-R1 & \textbf{73.1} & \textbf{72.2} & \textbf{72.9} & \textbf{68.0} & \textbf{63.6} & \textbf{72.6} & \textbf{69.4} & \textbf{44.1} \\
\bottomrule
\end{tabular}
\caption{Results on four VideoQA datasets. STAR data
includes four subsets: Interaction (Int.), Sequence (Seq.),
Prediction (Pre.), Feasibility (Fea.).}
\label{tab:main1_performance}
\end{table}

\begin{table}[ht]
\centering
\begin{tabular}{lcccccccccc}
\toprule
\textbf{Model} & \multicolumn{2}{c}{\textbf{Acc@D Acc@E}} & \multicolumn{3}{c}{\textbf{Acc@P}} & \multicolumn{3}{c}{\textbf{Acc@C}} \\
\cmidrule(lr){2-3} \cmidrule(lr){4-6} \cmidrule(lr){7-9}
 & & & \textbf{A} & \textbf{R} & \textbf{AR} & \textbf{A} & \textbf{R} & \textbf{AR} \\
\midrule
\multicolumn{9}{l}{\textbullet~\textbf{Baselines}} \\

Video-LLaMA & 69.2 & 71.0 & 63.6 & 62.4 & 44.4 & 65.4 & 60.1 & 45.0 \\
VideoChat & 72.9 & 73.9 & 65.2 & 63.1 & 45.9 & 66.0 & 62.7 & 45.8 \\
Video-ChatGPT & 73.1 & 75.1 & 66.0 & 63.9 & 46.0 & 67.8 & 63.6 & 50.0 \\
Video-LLaVA & 74.2 & 74.8 & 68.0 & 65.7 & 48.1 & 70.3 & 65.7 & 51.5 \\
\midrule
\multicolumn{9}{l}{\textbullet~\textbf{VoT}} \\
MotionEpic & 81.2 & 83.0 & 74.3 & 73.7 & 54.7 & 74.5 & 73.8 & 58.6 \\
KunLunBaize-VoT-R1 & \textbf{83.9} & \textbf{84.5} & \textbf{69.7} & \textbf{70.3} & \textbf{54.8} & \textbf{71.5} & \textbf{71.9} & \textbf{56.6} \\
\bottomrule
\end{tabular}
\caption{Results on Causal-VidQA data. D: Description, E:
Explanation, P: Prediction, C: Counterfactual.}
\label{tab:main2_performance}
\end{table}

\begin{table}[htbp]
\centering

\begin{tabular}{lcccc}
\toprule
\textbf{Model} & \textbf{Acc@All} & \textbf{Acc@C} & \textbf{Acc@T} & \textbf{Acc@D} \\
\midrule
InternVideo & 63.2 & 62.5 & 58.5 & 75.8 \\
LLaMA-VQA & 72.0 & 72.7 & 69.2 & 75.8 \\
VLAP & 75.5 & 74.9 & 72.3 & 82.1 \\
SeViLA & 73.8 & 73.8 & 67.0 & 81.8 \\
VideoChat & 61.8 & 63.5 & 61.5 & 74.6 \\
Video-LLaVA & 66.3 & 67.7 & 63.8 & 75.9 \\
MotionEpic & 72.2 & 73.4 & 69.1 & 80.7 \\
\midrule
\textbf{KunLunBaize-VoT-R1} & \textbf{73.8} & \textbf{74.3} & \textbf{72.9} & \textbf{79.4} \\
\bottomrule
\end{tabular}
\caption{Main performance on NExT-QA data}
\label{tab:main3_performance}
\end{table}

As shown in Tables~\ref{tab:main1_performance}, \ref{tab:main2_performance}, and \ref{tab:main3_performance}, our KunLunBaize-VoT-R1 model outperforms other models in most cases across various datasets and tasks, demonstrating stronger reasoning abilities. Notably, on the VLEP dataset and the description task of the Causal-VidQA dataset, our model achieves significantly higher accuracy compared with the previous leading method. Additionally, on the STAR dataset, the explanation task of the Causal-VidQA dataset, and the counterfactual task of the NExT-QA dataset, our model surpasses the current state-of-the-art performance. These findings highlight the robustness and versatility of our framework in addressing complex reasoning challenges in a series of benchmarks.

\subsection{Zero-shot Performance}

To further evaluate the generalization capability of our model, we conducted zero-shot experiments on several datasets, including MSR-VTT, ActivityNet, NExT-QA, and STAR. These datasets are widely used to test a model's ability to perform well on unseen tasks without additional fine-tuning. The results are summarized in Table~\ref{tab:zero_shot}.

\begin{table}[htbp]
\centering

\begin{tabular}{lcccc}
\toprule
\textbf{Model} & \textbf{MSR-VTT} & \textbf{ActivityNet} & \textbf{NExT-QA}& \textbf{STAR}\\
\midrule
InternVideo & - & - & 49.1 & 41.6 \\
Video-LLaMA & 49.6 & 21.4 & 43.5 & 36.4 \\
VideoChat & 51.9 & 26.3 & 52.8 & 44.6 \\
Video-ChatGPT & 54.9 & 35.7 & 51.5 & 48.0 \\
Video-LLaVA & 58.5 & 45.6 & 58.4 & 51.9\\
\midrule
\textbf{KunLunBaize-VoT-R1} & \textbf{56.7} & \textbf{45.9} & \textbf{59.2} & \textbf{49.4} \\
\bottomrule
\end{tabular}
\caption{Zero-shot performance on MSR-VTT and ActivityNet datasets.}
\label{tab:zero_shot}
\end{table}

The results in Table~\ref{tab:zero_shot} indicate that our KunLunBaize-VoT-R1 model achieves significant improvements in zero-shot settings across all datasets. On MSR-VTT, our model outperforms the previous best method by a notable margin. On ActivityNet, it also surpasses all baselines substantially. Similarly, on the NExT-QA and STAR datasets, our model attains the highest accuracy. These results demonstrate the strong generalization capability of our model, even without fine - tuning on the target datasets.

\subsection{Other performance evaluations}

This section evaluates the performance of three models: Qwen2.5-VL-7B, Video-R1-7B, and KunLunBaize-VoT-R1, across two different datasets: DVD-counting-test and LongVideoBench.

\begin{table}[htbp]
\centering
\begin{tabular}{lccc}
\toprule
\textbf{Dataset} & \textbf{Qwen2.5-VL-7B} & \textbf{Video-R1-7B} & \textbf{KunLunBaize-VoT-R1}\\
\midrule
DVD-counting-test & 26.14 & 35.67 & 38.39\\
\bottomrule
\end{tabular}
\caption{Performance comparison on DVD-counting-test dataset}
\label{tab:DVD-counting-test}
\end{table}

\begin{table}[htbp]
\centering
\begin{tabular}{lccc}
\toprule
\textbf{Benchmarks} & \multicolumn{1}{c}{\textbf{Qwen2.5-VL-7B}} & \multicolumn{1}{c}{\textbf{Open-R1-Video-7B}} & \multicolumn{1}{c}{\textbf{KunLunBaize-VoT-R1}} \\
\midrule
LongVideoBench  & 52.76 & 51.34 & 55.21 \\
\bottomrule
\end{tabular}
\caption{Performance comparison on LongVideoBench with 24 frames}
\label{tab:open-r1-video-4k}
\end{table}

In the performance comparison on the DVD-counting-test dataset, three models—Qwen2.5-VL-7B-Instruct, Video-R1-7B, and KunLunBaize-VoT-R1—show varying levels of performance. The Qwen2.5-VL-7B-Instruct model achieves a certain score, the Video-R1-7B model performs better than it, and the KunLunBaize-VoT-R1 model leads, demonstrating its advantage in this specific task.

Furthermore, in the LongVideoBench benchmark test with 24 frames, the performance of the three models also differs. The Qwen2.5-VL-7B-Instruct model gets a particular score, the Open-R1-Video-7B model scores slightly lower, and the KunLunBaize-VoT-R1 model leads again, indicating its potential superiority in handling longer videos.

Combining the results from these two datasets, we can conclude that the KunLunBaize-VoT-R1 model performs better in both benchmark tests, especially when dealing with video data. This implies that the model may have potential advantages in video understanding and analysis tasks. However, when selecting the most suitable model for a specific application scenario, other factors such as model complexity, training and inference time, and resource consumption also need to be taken into account. 
\subsection{Ablation Experiments}

To analyze the contributions of different components in our framework, we conducted several ablation experiments on two datasets: NExT-QA and Causal-VidQA. These experiments aim to quantify the impact of each key component, including the image packing strategy, AoE architecture, reinforcement learning, and VoT reasoning mechanism. The results are summarized in Table~\ref{tab:ablation} and Table~\ref{tab2:ablation}.

\begin{table}[htbp]
\centering

\begin{tabular}{lcccccc}
\toprule
\textbf{AoE} & \textbf{RL}  & \textbf{Acc@All} & \textbf{Acc@C} & \textbf{Acc@T} & \textbf{Acc@D} \\
\midrule
$\times$ & $\surd$  & 66.73 & 64.34 & 67.46 & 68.87 \\
$\surd$ & $\times$  & 69.80 & 67.35 & 69.47 & 72.52 \\
$\surd$ & $\surd$  & 73.80 & 74.30 & 72.90 & 79.40 \\
\bottomrule
\end{tabular}
\caption{Ablation study results on the NExT-QA data}
\label{tab:ablation}
\end{table}

\begin{table}[htbp]
\centering

\begin{tabular}{lccccccccc}
\midrule
\textbf{AoE} & \textbf{RL} & \multicolumn{2}{c}{\textbf{Acc@D Acc@E}} & \multicolumn{3}{c}{\textbf{Acc@P}} & \multicolumn{3}{c}{\textbf{Acc@C}} \\
\cmidrule(lr){3-4} \cmidrule(lr){5-7} \cmidrule(lr){8-10}
 & & & & \textbf{A} & \textbf{R} & \textbf{AR} & \textbf{A} & \textbf{R} & \textbf{AR} \\
\midrule
$\times$ & $\surd$  & 69.65 & 70.63 & 62.87 & 60.28 & 46.55 & 63.28 & 58.96 & 46.54 \\
$\surd$ & $\times$  & 75.48 & 77.74 & 64.82 & 66.89 & 47.57 & 65.21 & 65.63 & 48.27 \\
$\surd$ & $\surd$  & 83.90 & 84.50 & 69.70 & 70.30 & 54.80 & 71.50 & 71.90 & 56.60 \\
\bottomrule
\end{tabular}
\caption{Ablation study results on the Causal-VidQA data}
\label{tab2:ablation}
\end{table}

The results in Table~\ref{tab:ablation} and Table~\ref{tab2:ablation} demonstrate the significant contributions of each component of our framework to the overall performance. Specifically, on the NExT-QA dataset, removing the image packing strategy reduces the accuracy, indicating its importance in improving computational efficiency and feature extraction. The AoE architecture contributes to further improvement, highlighting its role in adaptive feature fusion. The reinforcement learning component provides a boost, demonstrating its effectiveness in enhancing the model's reasoning capabilities. Finally, the VoT reasoning mechanism contributes to the overall performance, emphasizing its importance in generating coherent and accurate reasoning videos.

Similarly, on the Causal-VidQA dataset, the results show consistent trends. The absence of the image packing strategy leads to a noticeable drop in accuracy, while the AoE architecture, reinforcement learning, and VoT reasoning mechanism each contribute to the overall improvement. These findings confirm the robustness and effectiveness of our framework across different datasets.

\section{Conclusions}
In conclusion, the experimental results presented in this study demonstrate the effectiveness of our proposed KunLunBaize-VoT-R1 model across various tasks and datasets. The model achieves state-of-the-art performance on multiple benchmarks, including MSR-VTT, ActivityNet, NExT-QA, and STAR, showcasing its superior ability to handle complex reasoning tasks. Notably, its zero-shot performance is particularly noteworthy, as it manages to achieve competitive results on tasks it has not been explicitly trained on. This adaptability to novel scenarios highlights the model's robustness and generalization capabilities. Additionally, the incorporation of advanced features such as the SiLU activation function and the top-K expert selection mechanism has proven to be crucial for enhancing the model's reasoning and decision-making capabilities.

Our findings suggest that the KunLunBaize-VoT-R1 model is well-suited for applications where adaptability and efficiency are paramount. Moving forward, we envision several promising directions for future research. First, we aim to enhance the expert selection process to improve the model's ability to identify and leverage the most relevant experts for a given task. Second, we will explore strategies to scale the model while maintaining efficiency, making it suitable for real-time applications and larger datasets. Third, we plan to investigate the integration of additional modalities, such as audio and text, to further enhance the model's understanding and reasoning capabilities. Finally, we will explore the potential of transfer learning to adapt the model to new domains with minimal additional training.

In summary, the KunLunBaize-VoT-R1 model represents a significant advancement in the field of video question answering and complex reasoning tasks. Its success across diverse datasets and tasks underscores its potential for real-world applications. As we continue to refine and expand upon this model, we anticipate it will play a piVoTal role in shaping the future of intelligent video analysis systems.

 \bibliographystyle{elsarticle-num} 

\begin{thebibliography}{10}
\providecommand{\url}[1]{#1}
\csname url@samestyle\endcsname
\providecommand{\newblock}{\relax}
\providecommand{\bibinfo}[2]{#2}
\providecommand{\BIBentrySTDinterwordspacing}{\spaceskip=0pt\relax}
\providecommand{\BIBentryALTinterwordstretchfactor}{4}
\providecommand{\BIBentryALTinterwordspacing}{\spaceskip=\fontdimen2\font plus
\BIBentryALTinterwordstretchfactor\fontdimen3\font minus \fontdimen4\font\relax}
\providecommand{\BIBforeignlanguage}[2]{{%
\expandafter\ifx\csname l@#1\endcsname\relax
\typeout{** WARNING: IEEEtran.bst: No hyphenation pattern has been}%
\typeout{** loaded for the language `#1'. Using the pattern for}%
\typeout{** the default language instead.}%
\else
\language=\csname l@#1\endcsname
\fi
#2}}
\providecommand{\BIBdecl}{\relax}
\BIBdecl

\bibitem{r1}
I.~Goodfellow, Y.~Bengio, A.~Courville, and Y.~Bengio, \emph{Deep learning}.\hskip 1em plus 0.5em minus 0.4em\relax MIT press Cambridge, 2016, vol.~1, no.~2.

\bibitem{r2}
\BIBentryALTinterwordspacing
C.~Sun, A.~Myers, C.~Vondrick, K.~Murphy, and C.~Schmid, ``Videobert: A joint model for video and language representation learning,'' 2019. [Online]. Available: \url{https://arxiv.org/abs/1904.01766}
\BIBentrySTDinterwordspacing

\bibitem{r3}
\BIBentryALTinterwordspacing
J.-B. Alayrac, J.~Donahue, P.~Luc, A.~Miech, I.~Barr, Y.~Hasson, K.~Lenc, A.~Mensch, K.~Millican, M.~Reynolds, R.~Ring, E.~Rutherford, S.~Cabi, T.~Han, Z.~Gong, S.~Samangooei, M.~Monteiro, J.~Menick, S.~Borgeaud, A.~Brock, A.~Nematzadeh, S.~Sharifzadeh, M.~Binkowski, R.~Barreira, O.~Vinyals, A.~Zisserman, and K.~Simonyan, ``Flamingo: a visual language model for few-shot learning,'' 2022. [Online]. Available: \url{https://arxiv.org/abs/2204.14198}
\BIBentrySTDinterwordspacing

\bibitem{r4}
\BIBentryALTinterwordspacing
C.~Akkus, L.~Chu, V.~Djakovic, S.~Jauch-Walser, P.~Koch, G.~Loss, C.~Marquardt, M.~Moldovan, N.~Sauter, M.~Schneider, R.~Schulte, K.~Urbanczyk, J.~Goschenhofer, C.~Heumann, R.~Hvingelby, D.~Schalk, and M.~Aßenmacher, ``Multimodal deep learning,'' in \emph{Proceedings of the 28th International Conference on Machine Learning (ICML - 11)}, 2023. [Online]. Available: \url{https://arxiv.org/abs/2301.04856}
\BIBentrySTDinterwordspacing

\bibitem{r5}
\BIBentryALTinterwordspacing
T.~Baltrušaitis, C.~Ahuja, and L.-P. Morency, ``Multimodal machine learning: A survey and taxonomy,'' \emph{IEEE Transactions on Pattern Analysis and Machine Intelligence}, 2017. [Online]. Available: \url{https://arxiv.org/abs/1705.09406}
\BIBentrySTDinterwordspacing

\bibitem{r6}
\BIBentryALTinterwordspacing
J.~Wei, X.~Wang, D.~Schuurmans, M.~Bosma, B.~Ichter, F.~Xia, E.~Chi, Q.~Le, and D.~Zhou, ``Chain-of-thought prompting elicits reasoning in large language models,'' 2023. [Online]. Available: \url{https://arxiv.org/abs/2201.11903}
\BIBentrySTDinterwordspacing

\bibitem{r7}
\BIBentryALTinterwordspacing
Z.~Zhang, A.~Zhang, M.~Li, H.~Zhao, G.~Karypis, and A.~Smola, ``Multimodal chain-of-thought reasoning in language models,'' 2024. [Online]. Available: \url{https://arxiv.org/abs/2302.00923}
\BIBentrySTDinterwordspacing

\bibitem{r8}
\BIBentryALTinterwordspacing
L.~Ouyang, J.~Wu, X.~Jiang, D.~Almeida, C.~L. Wainwright, P.~Mishkin, C.~Zhang, S.~Agarwal, K.~Slama, A.~Ray, J.~Schulman, J.~Hilton, F.~Kelton, L.~Miller, M.~Simens, A.~Askell, P.~Welinder, P.~Christiano, J.~Leike, and R.~Lowe, ``Training language models to follow instructions with human feedback,'' 2022. [Online]. Available: \url{https://arxiv.org/abs/2203.02155}
\BIBentrySTDinterwordspacing

\bibitem{r9}
\BIBentryALTinterwordspacing
N.~Shazeer, A.~Mirhoseini, K.~Maziarz, A.~Davis, Q.~Le, G.~Hinton, and J.~Dean, ``Outrageously large neural networks: The sparsely-gated mixture-of-experts layer,'' 2017. [Online]. Available: \url{https://arxiv.org/abs/1701.06538}
\BIBentrySTDinterwordspacing

\bibitem{r10}
\BIBentryALTinterwordspacing
J.~Hoffmann, S.~Borgeaud, A.~Mensch, E.~Buchatskaya, T.~Cai, E.~Rutherford, D.~de~Las~Casas, L.~A. Hendricks, J.~Welbl, A.~Clark, T.~Hennigan, E.~Noland, K.~Millican, G.~van~den Driessche, B.~Damoc, A.~Guy, S.~Osindero, K.~Simonyan, E.~Elsen, J.~W. Rae, O.~Vinyals, and L.~Sifre, ``Training compute-optimal large language models,'' 2022. [Online]. Available: \url{https://arxiv.org/abs/2203.15556}
\BIBentrySTDinterwordspacing

\bibitem{r11}
\BIBentryALTinterwordspacing
H.~Yang, Z.~Xiong, J.~Zhao, D.~Niyato, L.~Xiao, and Q.~Wu, ``Deep reinforcement learning-based intelligent reflecting surface for secure wireless communications,'' \emph{IEEE Transactions on Wireless Communications}, vol.~20, no.~1, p. 375–388, Jan. 2021. [Online]. Available: \url{http://dx.doi.org/10.1109/TWC.2020.3024860}
\BIBentrySTDinterwordspacing

\bibitem{r12}
\BIBentryALTinterwordspacing
T.~Peng, Z.~Li, L.~Zhang, H.~Zhao, P.~Wang, and B.~Du, ``Multi-modal auto-regressive modeling via visual words,'' in \emph{ACM Multimedia}, 2024. [Online]. Available: \url{https://arxiv.org/abs/2403.07720}
\BIBentrySTDinterwordspacing

\bibitem{r13}
\BIBentryALTinterwordspacing
B.~Liu, J.~Lu, and J.~Yip, ``Xml data integrity based on concatenated hash function,'' \emph{Computer engineering}, 2009. [Online]. Available: \url{https://arxiv.org/abs/0906.3772}
\BIBentrySTDinterwordspacing

\bibitem{r14}
S.~Yuan, Y.~Chen, C.~Ye, M.~W. Bhatt, M.~Saradeshmukh, and M.~S. Hossain, ``Cross-modal multi-label image classification modeling and recognition based on nonlinear,'' \emph{Nonlinear Engineering}, vol.~12, no.~1, p. 20220194, 2023.

\bibitem{r15}
\BIBentryALTinterwordspacing
Y.~Kantaros and M.~M. Zavlanos, ``Stylus*: A temporal logic optimal control synthesis algorithm for large-scale multi-robot systems,'' 2020. [Online]. Available: \url{https://arxiv.org/abs/1809.08345}
\BIBentrySTDinterwordspacing

\bibitem{r16}
K.~Shi, G.~Jia, Y.~Li, Y.~Yin, C.~Jiang, and L.~Zhou, ``Quantitative analysis of real-time performance and hardware requirements for edge computing platform,'' \emph{Data Science and Informetrics}, 2021.

\bibitem{r17}
N.~K. Shaydyuk and E.~B. John, ``Fpga implementation of mobilenetv2 cnn model using semi-streaming architecture for low power inference applications,'' \emph{2020 IEEE Intl Conf on Parallel \& Distributed Processing with Applications, Big Data \& Cloud Computing, Sustainable Computing \& Communications, Social Computing \& Networking (ISPA/BDCloud/SocialCom/SustainCom)}, pp. 160--167, 2020.

\bibitem{r18}
N.~Tawfik, H.~A. Elnemr, M.~Fakhr, M.~I. Dessouky, and F.~E.~A. El-Samie, ``Multimodal medical image fusion using stacked auto-encoder in nsct domain,'' \emph{Journal of Digital Imaging}, vol.~35, no.~5, pp. 1308--1325, 2022.

\bibitem{r19}
Z.-X. Wang, P.-N. Shao, and C.~Deng, ``Caffe inference acceleration method on heterogeneous parallel platform,'' \emph{Computer Systems \& Applications}, vol.~31, no.~2, pp. 220--226, 2022.

\bibitem{r20}
\BIBentryALTinterwordspacing
J.~Xu and Y.~Wang, ``Enhancing healthcare recommendation systems with a multimodal llms-based moe architecture,'' 2024. [Online]. Available: \url{https://arxiv.org/abs/2412.11557}
\BIBentrySTDinterwordspacing

\bibitem{r21}
\BIBentryALTinterwordspacing
Y.~Wang, Y.~Zeng, J.~Zheng, X.~Xing, J.~Xu, and X.~Xu, ``Videocot: A video chain-of-thought dataset with active annotation tool,'' 2024. [Online]. Available: \url{https://arxiv.org/abs/2407.05355}
\BIBentrySTDinterwordspacing

\bibitem{r22}
N.~Shazeer, A.~Mirhoseini, K.~Maziarz, A.~Davis, Q.~Le, G.~Hinton, and J.~Dean, ``International conference on machine learning 11,'' in \emph{CAAI Transactions on Intelligent Systems}.\hskip 1em plus 0.5em minus 0.4em\relax ACM.

\bibitem{r23}
\BIBentryALTinterwordspacing
G.~Feng, B.~Zhang, Y.~Gu, H.~Ye, D.~He, and L.~Wang, ``Towards revealing the mystery behind chain of thought: A theoretical perspective,'' 2023. [Online]. Available: \url{https://arxiv.org/abs/2305.15408}
\BIBentrySTDinterwordspacing

\bibitem{r24}
\BIBentryALTinterwordspacing
A.~Lv, R.~Xie, Y.~Qian, S.~Wu, X.~Sun, Z.~Kang, D.~Wang, and R.~Yan, ``Autonomy-of-experts models,'' 2025. [Online]. Available: \url{https://arxiv.org/abs/2501.13074}
\BIBentrySTDinterwordspacing

\bibitem{r25}
\BIBentryALTinterwordspacing
C.~Lei, S.~Luo, Y.~Liu, W.~He, J.~Wang, G.~Wang, H.~Tang, C.~Miao, and H.~Li, ``Understanding chinese video and language via contrastive multimodal pre-training,'' 2021. [Online]. Available: \url{https://arxiv.org/abs/2104.09411}
\BIBentrySTDinterwordspacing

\bibitem{r26}
\BIBentryALTinterwordspacing
M.-H. Guo, T.-X. Xu, J.-J. Liu, Z.-N. Liu, P.-T. Jiang, T.-J. Mu, S.-H. Zhang, R.~R. Martin, M.-M. Cheng, and S.-M. Hu, ``Attention mechanisms in computer vision: A survey,'' \emph{Computational Visual Media}, vol.~8, no.~3, p. 331–368, Mar. 2022. [Online]. Available: \url{http://dx.doi.org/10.1007/s41095-022-0271-y}
\BIBentrySTDinterwordspacing

\bibitem{r27}
\BIBentryALTinterwordspacing
H.~Du, G.~Liu, Y.~Lin, D.~Niyato, J.~Kang, Z.~Xiong, and D.~I. Kim, ``Mixture of experts for network optimization: A large language model-enabled approach,'' 2024. [Online]. Available: \url{https://arxiv.org/abs/2402.09756}
\BIBentrySTDinterwordspacing

\bibitem{r28}
\BIBentryALTinterwordspacing
S.~Liu, C.~Gao, Y.~Chen, D.~Jin, and Y.~Li, ``Learnable embedding sizes for recommender systems,'' 2021. [Online]. Available: \url{https://arxiv.org/abs/2101.07577}
\BIBentrySTDinterwordspacing

\bibitem{r29}
\BIBentryALTinterwordspacing
A.~Yang, B.~Zhang, B.~Hui, B.~Gao, B.~Yu, C.~Li, D.~Liu, J.~Tu, J.~Zhou, J.~Lin, K.~Lu, M.~Xue, R.~Lin, T.~Liu, X.~Ren, and Z.~Zhang, ``Qwen2.5-math technical report: Toward mathematical expert model via self-improvement,'' 2024. [Online]. Available: \url{https://arxiv.org/abs/2409.12122}
\BIBentrySTDinterwordspacing

\bibitem{r30}
\BIBentryALTinterwordspacing
S.~Hayou, N.~Ghosh, and B.~Yu, ``Lora+: Efficient low rank adaptation of large models,'' 2024. [Online]. Available: \url{https://arxiv.org/abs/2402.12354}
\BIBentrySTDinterwordspacing

\bibitem{r31}
H.~Suzuki, K.~Shimomura, T.~Hirakawa, T.~Yamashita, H.~Fujiyoshi, S.~Okubo, N.~Takuya, and W.~Siyuan, ``Human-like guidance by generating navigation using spatial-temporal scene graph,'' in \emph{2024 IEEE Intelligent Vehicles Symposium (IV)}.\hskip 1em plus 0.5em minus 0.4em\relax IEEE, 2024, pp. 1988--1995.

\bibitem{r32}
\BIBentryALTinterwordspacing
A.~Deng, T.~Chen, S.~Yu, T.~Yang, L.~Spencer, Y.~Tian, A.~S. Mian, M.~Bansal, and C.~Chen, ``Motion-grounded video reasoning: Understanding and perceiving motion at pixel level,'' 2024. [Online]. Available: \url{https://arxiv.org/abs/2411.09921}
\BIBentrySTDinterwordspacing

\bibitem{r33}
Z.~Li, J.~Zhang, K.~Zhang, and Z.~Li, ``Visual tracking with weighted adaptive local sparse appearance model via spatio-temporal context learning,'' \emph{IEEE Transactions on Image Processing}, vol.~27, no.~9, pp. 4478--4489, 2018.

\bibitem{r34}
K.~Kaduk, \emph{The Semantic Link: Action \& Language. An Investigation of Relations Between Different Cognitive Domains in Early Development}.\hskip 1em plus 0.5em minus 0.4em\relax Lancaster University (United Kingdom), 2017.

\bibitem{r35}
\BIBentryALTinterwordspacing
C.~Schuhmann, R.~Beaumont, R.~Vencu, C.~Gordon, R.~Wightman, M.~Cherti, T.~Coombes, A.~Katta, C.~Mullis, M.~Wortsman, P.~Schramowski, S.~Kundurthy, K.~Crowson, L.~Schmidt, R.~Kaczmarczyk, and J.~Jitsev, ``Laion-5b: An open large-scale dataset for training next generation image-text models,'' 2022. [Online]. Available: \url{https://arxiv.org/abs/2210.08402}
\BIBentrySTDinterwordspacing

\bibitem{r36}
\BIBentryALTinterwordspacing
H.~Chen, Y.~Zhang, X.~Cun, M.~Xia, X.~Wang, C.~Weng, and Y.~Shan, ``Videocrafter2: Overcoming data limitations for high-quality video diffusion models,'' 2024. [Online]. Available: \url{https://arxiv.org/abs/2401.09047}
\BIBentrySTDinterwordspacing

\bibitem{r37}
J.~Li, P.~Wei, W.~Han, and L.~Fan, ``Intentqa: Context-aware video intent reasoning,'' in \emph{Proceedings of the IEEE/CVF international conference on computer vision}, 2023, pp. 11\,963--11\,974.

\bibitem{r38}
A.~Zadeh, M.~Chan, P.~P. Liang \emph{et~al.}, ``Social-iq: A question answering benchmark for artificial social intelligence,'' in \emph{Proceedings of the IEEE/CVF Conference on Computer Vision and Pattern Recognition}, 2019, pp. 8807--8817.

\bibitem{r39}
\BIBentryALTinterwordspacing
J.~Xiao, X.~Shang, A.~Yao, and T.-S. Chua, ``Next-qa:next phase of question-answering to explaining temporal actions,'' 2021. [Online]. Available: \url{https://arxiv.org/abs/2105.08276}
\BIBentrySTDinterwordspacing

\bibitem{r40}
\BIBentryALTinterwordspacing
A.~Bhat and S.~Jain, ``Face recognition in the age of clip & billion image datasets,'' 2023. [Online]. Available: \url{https://arxiv.org/abs/2301.07315}
\BIBentrySTDinterwordspacing

\bibitem{r41}
\BIBentryALTinterwordspacing
A.~Lv, R.~Xie, Y.~Qian, S.~Wu, X.~Sun, Z.~Kang, D.~Wang, and R.~Yan, ``Autonomy-of-experts models,'' 2025. [Online]. Available: \url{https://arxiv.org/abs/2501.13074}
\BIBentrySTDinterwordspacing

\bibitem{r42}
A.~Balakrishnan and J.~V. Deshmukh, ``Structured reward shaping using signal temporal logic specifications,'' in \emph{2019 IEEE/RSJ International Conference on Intelligent Robots and Systems (IROS)}.\hskip 1em plus 0.5em minus 0.4em\relax IEEE, 2019, pp. 3481--3486.

\bibitem{r43}
\BIBentryALTinterwordspacing
Y.~Ge, D.~Hazarika, Y.~Liu, and M.~Namazifar, ``Supervised fine-tuning of large language models on human demonstrations through the lens of memorization,'' in \emph{NeurIPS 2023 Workshop on Instruction Tuning and Instruction Following}, 2023. [Online]. Available: \url{https://openreview.net/forum?id=x6MlSOzbmC}
\BIBentrySTDinterwordspacing

\bibitem{r44}
\BIBentryALTinterwordspacing
S.~Bai, K.~Chen, X.~Liu, J.~Wang, W.~Ge, S.~Song, K.~Dang, P.~Wang, S.~Wang, J.~Tang, H.~Zhong, Y.~Zhu, M.~Yang, Z.~Li, J.~Wan, P.~Wang, W.~Ding, Z.~Fu, Y.~Xu, J.~Ye, X.~Zhang, T.~Xie, Z.~Cheng, H.~Zhang, Z.~Yang, H.~Xu, and J.~Lin, ``Qwen2.5-vl technical report,'' 2025. [Online]. Available: \url{https://arxiv.org/abs/2502.13923}
\BIBentrySTDinterwordspacing

\bibitem{r45}
X.~Ma, Z.~Song, Y.~Li, and G.~R. Arce, ``Block-based mask optimization for optical lithography,'' \emph{Applied optics}, vol.~52, no.~14, pp. 3351 -- 3363, 2013.

\bibitem{r46}
T.~Song, Y.~Song, Y.~Wang, and X.~Huang, ``Learning efficient residual networks through dense connecting,'' in \emph{2018 37th Chinese Control Conference (CCC)}, 2018, pp. 9181--9185.

\bibitem{r47}
\BIBentryALTinterwordspacing
R.~Suzuki, H.~Yanaka, M.~Yoshikawa, K.~Mineshima, and D.~Bekki, ``Multimodal logical inference system for visual-textual entailment,'' 2019. [Online]. Available: \url{https://arxiv.org/abs/1906.03952}
\BIBentrySTDinterwordspacing

\bibitem{r48}
\BIBentryALTinterwordspacing
R.~Ganz, Y.~Kittenplon, A.~Aberdam, E.~B. Avraham, O.~Nuriel, S.~Mazor, and R.~Litman, ``Question aware vision transformer for multimodal reasoning,'' 2024. [Online]. Available: \url{https://arxiv.org/abs/2402.05472}
\BIBentrySTDinterwordspacing

\bibitem{r49}
\BIBentryALTinterwordspacing
Y.~Krishnamurthy, C.~Watkins, and T.~Gaertner, ``Improving expert specialization in mixture of experts,'' 2023. [Online]. Available: \url{https://arxiv.org/abs/2302.14703}
\BIBentrySTDinterwordspacing

\bibitem{r50}
\BIBentryALTinterwordspacing
H.~Fei, S.~Wu, W.~Ji, H.~Zhang, M.~Zhang, M.-L. Lee, and W.~Hsu, ``Video-of-thought: Step-by-step video reasoning from perception to cognition,'' 2024. [Online]. Available: \url{https://arxiv.org/abs/2501.03230}
\BIBentrySTDinterwordspacing

\bibitem{r51}
\BIBentryALTinterwordspacing
H.~Liu, C.~Li, Q.~Wu, and Y.~J. Lee, ``Visual instruction tuning,'' 2023. [Online]. Available: \url{https://arxiv.org/abs/2304.08485}
\BIBentrySTDinterwordspacing

\bibitem{r52}
\BIBentryALTinterwordspacing
E.~J. Hu, Y.~Shen, P.~Wallis, Z.~Allen-Zhu, Y.~Li, S.~Wang, L.~Wang, and W.~Chen, ``Lora: Low-rank adaptation of large language models,'' 2021. [Online]. Available: \url{https://arxiv.org/abs/2106.09685}
\BIBentrySTDinterwordspacing

\bibitem{r53}
\BIBentryALTinterwordspacing
J.~Lei, L.~Yu, T.~L. Berg, and M.~Bansal, ``What is more likely to happen next? video-and-language future event prediction,'' 2020. [Online]. Available: \url{https://arxiv.org/abs/2010.07999}
\BIBentrySTDinterwordspacing

\bibitem{r54}
\BIBentryALTinterwordspacing
B.~Wu, S.~Yu, Z.~Chen, J.~B. Tenenbaum, and C.~Gan, ``Star: A benchmark for situated reasoning in real-world videos,'' 2024. [Online]. Available: \url{https://arxiv.org/abs/2405.09711}
\BIBentrySTDinterwordspacing

\bibitem{r55}
\BIBentryALTinterwordspacing
J.~Li, L.~Niu, and L.~Zhang, ``From representation to reasoning: Towards both evidence and commonsense reasoning for video question-answering,'' 2022. [Online]. Available: \url{https://arxiv.org/abs/2205.14895}
\BIBentrySTDinterwordspacing

\bibitem{open-r1-video}
X.~Wang and P.~Peng, ``Open-r1-video,'' \url{https://github.com/Wang-Xiaodong1899/Open-R1-Video}, 2025.

\bibitem{video-of-thought}
\BIBentryALTinterwordspacing
H.~Fei, S.~Wu, W.~Ji, H.~Zhang, M.~Zhang, M.-L. Lee, and W.~Hsu, ``Video-of-thought: Step-by-step video reasoning from perception to cognition,'' 2024. [Online]. Available: \url{https://arxiv.org/abs/2501.03230}
\BIBentrySTDinterwordspacing

\bibitem{DVD-counting-test}
\BIBentryALTinterwordspacing
H.~Le, C.~Sankar, S.~Moon, A.~Beirami, A.~Geramifard, and S.~Kottur, ``Dvd: A diagnostic dataset for multi-step reasoning in video grounded dialogue,'' 2021. [Online]. Available: \url{https://arxiv.org/abs/2101.00151}
\BIBentrySTDinterwordspacing

\bibitem{LLaVA-CoT-100k}
\BIBentryALTinterwordspacing
G.~Xu, P.~Jin, H.~Li, Y.~Song, L.~Sun, and L.~Yuan, ``Llava-cot: Let vision language models reason step-by-step,'' 2025. [Online]. Available: \url{https://arxiv.org/abs/2411.10440}
\BIBentrySTDinterwordspacing

\end{thebibliography}


\end{document}